%% file: root.tex
\documentclass[letterpaper, 10 pt, conference]{ieeeconf}  

\IEEEoverridecommandlockouts                              

\usepackage{graphics} 
\usepackage{amsmath} 
\usepackage{amssymb}  
\usepackage{xcolor}
\usepackage[hidelinks]{hyperref}
\usepackage{graphicx}
\usepackage{wasysym}
\usepackage{ulem} 
\usepackage{tcolorbox}
\usepackage{booktabs} 
\usepackage{siunitx}
\usepackage{textcomp}

\input{paper_style.sty}

\newcommand\copyrighttext{%
    \footnotesize 
    \color{gray}
    \textcopyright  
    2025
    \ IEEE. Personal use of this material is permitted.  Permission from IEEE must be obtained for all other uses, in any current or future media, including reprinting/republishing this material for advertising or promotional purposes, creating new collective works, for resale or redistribution to servers or lists, or reuse of any copyrighted component of this work in other works.
}

\title{\LARGE \bf
Toward Dynamic Control of Tendon-driven Continuum Robots\\using Clarke Transform
}

\author{Christian Muhmann$^{1,2}$, Reinhard M.~Grassmann$^{1, 3}$, Max Bartholdt$^{2}$, and Jessica Burgner-Kahrs$^{1, 3}$
 \thanks{We acknowledge the support of the Natural Sciences and Engineering Research Council of Canada (NSERC), [RGPIN-2019-04846].}
 \thanks{$^{1}$Continuum Robotics Laboratory, Department of Mathematical and Computational Sciences, University of Toronto, Mississauga, ON L5L 1C6, Canada {\tt\small reinhard.grassmann@utoronto.ca}}
 \thanks{$^{2}$Institute of Mechatronic Systems, Faculty of Mechanical Engineering, Leibniz University Hannover, 30823 Garbsen, Germany {\tt\small christian.muhmann@alumni-uni-hannover.de}}
 \thanks{$^{3}$University of Toronto Robotics Institute, Toronto, M5S 1A4, Canada}
}

\begin{document}

\maketitle
\thispagestyle{empty}
\pagestyle{empty}

\begin{abstract}

\input{Abstract/abstract}

\end{abstract}

\begin{tikzpicture}[remember picture,overlay]
        \node[anchor=south,yshift=10pt] at (current page.south) {\parbox{\dimexpr0.75\textwidth-\fboxsep-\fboxrule\relax}{\copyrighttext}};
\end{tikzpicture}


\input{Introduction/introduction}

\input{Dynamics/dynamics}

\input{Control_Strategies/control_strategies}

\input{Validation/validation}

\input{Discussion/discussion}

\input{Conclusion/conclusion}


\section*{ACKNOWLEDGMENT}
The first author was an \ac{IVGS} at the University of Toronto from the Leibniz University Hannover, Germany. We thank Thomas Seel for supporting the \ac{IVGS} program.

\addcontentsline{toc}{section}{REFERENCES}
\bibliographystyle{IEEEtran}
\bibliography{IEEEabrv, literature}

\end{document}

%% file: Abstract/abstract.tex
In this paper, we propose a dynamic model and control framework for \acp{TDCR} with multiple segments and an arbitrary number of tendons per segment.
Our approach leverages the Clarke transform, the Euler-Lagrange formalism, and the \acl{CC} assumption to formulate a dynamic model on a two-dimensional manifold embedded in the joint space that inherently satisfies tendon constraints.
We present linear and constraint-informed controllers that operate directly on this manifold, along with practical methods for preventing negative tendon forces without compromising control fidelity.
This opens up new design possibilities for overactuated \acp{TDCR} with improved force distribution and stiffness without increasing controller complexity.
We validate these approaches in simulation and on a physical prototype with one segment and five tendons, demonstrating accurate dynamic behavior and robust trajectory tracking under real-time conditions.

%% file: Introduction/introduction.tex
\section{Introduction}
\label{sec:introduction}

\begin{figure}[tbp]
    \vspace*{0.75em}
    \centering
    \includegraphics[width=0.65\columnwidth]{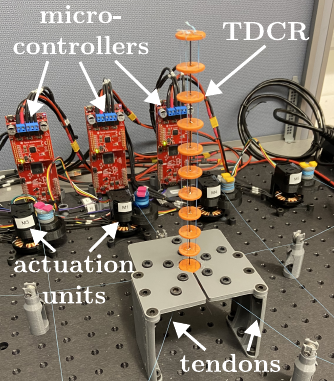}
    \vspace{-1em}
    \caption{\ac{TDCR} prototype with one segment and five tendons. Each tendon is driven by a backdrivable actuation unit, which is controlled by a micro-controller~\cite{Grassmann2024}. The robot is controlled by linear controllers as described in this work.}
    \label{fig:test_bench}
    \vspace{-1.75em}
\end{figure}

A \acf{TDCR} consists of an elastic backbone with tendons routed along its length to transmit actuation forces from an external unit to the compliant structure~\cite{BurgnerKahrs2015}.
Spacer disks guide the tendons along the backbone, terminating at end disks to define segments as shown in Fig.~\ref{fig:test_bench}.
Pulling the tendons generates bending motions, with the segment tip moving in a curved plane with two \ac{DoF}~\cite{Grassmann2022}.
Stacking multiple segments extends the robot's positioning capabilities beyond this plane.

Most \ac{TDCR} designs are limited to four tendons per segment, as this allows simplifications of tendon constraints and forward robot-dependent kinematics~\cite{Rao2021, Webster2010}.
However, using more tendons reduces the maximum tendon load~\cite{Dalvand2018}, improving force distribution, absorption, and delivery.
Additional tendons also increase stiffness~\cite{Butler2019} and enhance safety due to redundancy.

\subsection{Related Work}
\label{subsec:related_work}

Dynamic models are essential, as they provide a way to simulate robots as well as to develop robot designs and control schemes.
For \acp{TDCR}, Cosserat rod models are widely used in dynamic modeling due to their accuracy in large deformation analyses \cite{Jalali2022}.
These models, which couple the backbone and tendons, result in a set of \acp{PDE}.
However, restrictive time step conditions limit real-time applications in simulation and control~\cite{Rucker2011}.
Therefore, maintaining computational efficiency while preserving model realism is challenging \cite{Armanini2023}.

A numerical framework for solving Cosserat-based forward dynamics in \acp{CR} is introduced in~\cite{Till2019}, where implicit differentiation transforms the \acp{PDE} into a set of \acp{ODE} in the spatial domain, solved via numerical integration and shooting methods. 
While applicable to various \acp{CR}, including \acp{TDCR}~\cite{JanabiSharifi2021}, experimental validation only demonstrates soft real-time performance~\cite{Till2019} in systems with a limited number of \ac{DoF}.
To improve computational efficiency,~\cite{Renda2018} proposes a \acl{PCS} discretization approach, accounting for shear and torsional deformations, which are crucial for handling out-of-plane external loads.
This is extended in~\cite{Renda2020} to a variable-strain model, formulating the dynamics as a minimal set of \acp{ODE} in matrix form.
Although strain-based models are particularly suitable for control purposes, Cosserat-based models require extensive modeling states and remain computationally expensive~\cite{Jalali2022}.

An inverse dynamic model for a pneumatically actuated \ac{CR} with multiple sections is derived by \cite{Falkenhahn2014} using the Euler-Lagrange formalism and the \ac{CC} assumption.
A similar approach is implemented for inverse dynamics of \acp{TDCR} in arc space, both for one-segment~\cite{He2013} and multi-segment robots~\cite{Aner2024}.
These models leverage symbolic precomputation of the dynamic equations, ensuring computational efficiency for real-time simulations and control.
However, most used arc space parametrization introduces kinematic singularities~\cite{Santina2020}.
To address singularities in arc space,~\cite{Santina2020} introduces an improved parametrization for a \ac{CR} with four actuators. 
Similarly,~\cite{Qu2016,Allen2020,Dian2022} propose improved representations for robots with three or four actuators per segment.

In~\cite{GrassmannSenykBurgner-Kahrs_submitted_2024}, the Clarke transform is introduced to disentangle tendon constraints and map high-dimensional joint space representations onto an embedded two-dimensional manifold.
It offers the potential to develop approaches directly on the manifold that ensure intrinsic tendon constraint compliance, \textit{i.e.}, the sum of all tendon displacements is zero.
The derived Clarke coordinates~\cite{GrassmannSenykBurgner-Kahrs_submitted_2024} unify previous parametrizations ~\cite{Santina2020,Qu2016,Allen2020,Dian2022} and extend their applicability to robots with $n \geq 3$ symmetrically arranged tendons.
This generalization offers a robot-type-agnostic framework that integrates the benefits of previous methods and applies to various \acp{CR}, including \acp{TDCR}, enhancing the consistency and generality of these approaches.

\subsection{Contribution}
\label{subsec:contribution}

This work addresses the limited exploration of \ac{TDCR} designs and control strategies using a larger number of actuators per segment, which has been limited by the lack of analytical solutions for forward robot-dependent mappings.
We propose a computationally efficient dynamic model for \acp{TDCR} with multiple segments and $n$ tendons per segment, based on the Euler-Lagrange formalism and the \ac{CC} assumption.
By leveraging the Clarke transform \cite{GrassmannSenykBurgner-Kahrs_submitted_2024}, the dynamic model is reformulated on a \SI{2}{\ac{DoF}} manifold embedded in the joint space, simplifying its representation and enabling efficient simulations.
Additionally, the Clarke transform allows for constraint-informed controllers and enables the control of an entire segment using only two control parameters.
We utilize linear controllers, validating them in simulation and on a physical \ac{TDCR} with one segment and five tendons, as shown in Fig.~\ref{fig:test_bench}.

In particular, the main contributions are:
\begin{itemize}
    \item utilizing the Clarke transform in a generalized dynamic model of a \ac{TDCR} with multiple segments and $n$ tendons per segment based on the Euler-Lagrange formalism and the \ac{CC} assumption,
    \item implementation of linear PID and PD control for a \ac{TDCR} with $n$ tendons per segment on the \SI{2}{\ac{DoF}} manifold within the joint space,
    \item validation of the dynamic model and control schemes through simulations and experiments.
\end{itemize}

%% file: Dynamics/dynamics.tex
\section{Dynamic model}
\label{sec:dynamic_model}

In the following, we provide a self-contained description of the dynamic model using the Euler-Lagrange formalism and Clarke transform.
For this, we assume \ac{CC}~\cite{Webster2010} and fully constrained tendon path~\cite{Rao2021}.

\subsection{Kinematic Model}
\label{subsec:kinematic_model}

\begin{figure}
    \vspace*{0.75em}
    \centering
    \includegraphics[width=\columnwidth]{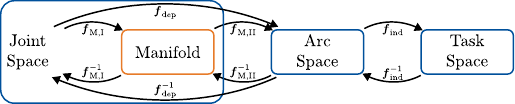}
    \vspace{-1.5em}
    \caption{
    Mappings between kinematic spaces of a \ac{TDCR}.
    The robot-independent mapping is denoted by $\boldsymbol{f}_{\mathrm{ind}}$, while the robot-dependent mapping is given by $\boldsymbol{f}_{\mathrm{dep}}=\boldsymbol{f}_{\mathrm{M,I}} \circ \boldsymbol{f}_{\mathrm{M,II}}$.
    The mappings from joint space to its manifold and from manifold to arc space are represented by $\boldsymbol{f}_{\mathrm{M,I}}$ and $\boldsymbol{f}_{\mathrm{M,II}}$, respectively.
    All inverse mappings are indicated by $\boldsymbol{f}^{-1}$.
    }
    \label{fig:scheme_mapping_new}
\end{figure}

As shown in Fig.~\ref{fig:scheme_mapping_new}, the kinematics of a \ac{TDCR} is divided into two mappings: the robot-independent mapping $\boldsymbol{f}_{\mathrm{ind}}$ describing the relationship between arc parameters and task-space coordinates, and the robot-dependent mapping $\boldsymbol{f}_{\mathrm{dep}}$ capturing the transformation between tendon displacements and arc parameters.
Note that the arc space is often referred to as configuration space~\cite{Webster2010}, and describes the bending angle and direction used in \ac{CC} models.
The robot-independent mapping $\boldsymbol{f}_{\mathrm{ind}}$ is well-established in the literature and has been addressed using various approaches~\cite{Webster2010, Rao2021}, whereas the approach proposed in~\cite{GrassmannSenykBurgner-Kahrs_submitted_2024} provides a generalized closed-form solution for the robot-dependent mapping $\boldsymbol{f}_{\mathrm{dep}}$.

Consider a \ac{TDCR} with $m$ segments and $n$ tendons per segment where the joint space of segment $i$ is defined by the tendon displacements $\boldsymbol{q}_i=[q_{i1},q_{i2},\dots,q_{in}]\transpose\in \mathbb{R}^n$, subject to the tendon constraint
\begin{align}
    \sum_{j=1}^{n} q_{ij} = 0.
    \label{eq:tendon_constraint}
\end{align}
The tendons run through guide holes arranged on the spacer disks, which are evenly distributed on a circle with radius $r_{\mathrm{d}}$.
Note that we assume no slack in the tendons.
The Clarke transform disentangles this constraint, providing a linear mapping between the joint space and a \SI{2}{DoF} manifold embedded within it.
The two variables $\boldsymbol{q}_{\mathrm{M}_i}=[q_{\mathrm{Re},i},q_{\mathrm{Im},i}]\transpose\in \mathbb{R}^2$ of the manifold are called Clarke coordinates \cite{GrassmannSenykBurgner-Kahrs_submitted_2024}.

This forward mapping $\boldsymbol{f}_{\mathrm{M,I}}$ can be expressed as
\begin{equation}
    \begin{bmatrix}
    q_{\mathrm{Re},i} \\
    q_{\mathrm{Im},i}
    \end{bmatrix}=
    \MP\boldsymbol{q}_{i},
    \label{eq:FM_JointSpace2Manifold}
\end{equation}
with a generalized Clarke transformation matrix~\cite{GrassmannSenykBurgner-Kahrs_submitted_2024}
\begin{equation}
    \hspace*{-0.75em}
    \MP
    =
    \!
    \dfrac{2}{n}
    \!
    \begin{bmatrix}
    \cos\!\left({0}\right) & \hspace{-0.25em}\cos\!\left({2\pi\dfrac{1}{n}}\right) & \hspace{-0.25em}\dots & \hspace{-0.25em}\cos\!\left({2\pi\dfrac{n-1}{n}}\right) \\
    \sin\!\left({0}\right) & \hspace{-0.25em}\sin\!\left({2\pi\dfrac{1}{n}}\right) & \hspace{-0.25em}\dots & \hspace{-0.25em}\sin\!\left({2\pi\dfrac{n-1}{n}}\right)
    \end{bmatrix}.
\end{equation}
Moreover, the inverse mapping $\boldsymbol{f}_{\mathrm{M,I}}^{-1}$ results in
\begin{equation}
    \boldsymbol{q}_{i} =
    \MPinv
    \begin{bmatrix}
    q_{\mathrm{Re},i} \\
    q_{\mathrm{Im},i}
    \end{bmatrix},
    \label{eq:IM_JointSpace2Manifold}
\end{equation}
with a generalized inverse Clarke transformation matrix~\cite{GrassmannSenykBurgner-Kahrs_submitted_2024}
\begin{equation}
    \MPinv = 
    \begin{bmatrix}
    \cos\left({0}\right) & \sin\left({0}\right) \\
    \cos\left({2\pi\dfrac{1}{n}}\right) & \sin\left({2\pi\dfrac{1}{n}}\right) \\
    \vdots & \vdots \\
    \cos\left({2\pi\dfrac{n-1}{n}}\right)
     & \sin\left({2\pi\dfrac{n-1}{n}}\right)
    \end{bmatrix}.
\end{equation}
The notation for the inverse Clarke transformation matrix follows~\cite{GrassmannSenykBurgner-Kahrs_submitted_2024}, where $()^{-1}$ refers to the inverse transformation, not the matrix inverse of $\MP$.

\begin{figure}
    \vspace*{0.75em}
    \centering
    \includegraphics[width=0.5\linewidth]{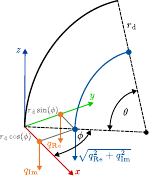}
    \caption{
    Physical interpretation of the Clarke coordinates.
    The blue line lies within the bending plane. The length difference to the arc length is the virtual displacement.
    The virtual displacement~\cite{GrassmannSenykBurgner-Kahrs_submitted_2024} is equal to $\sqrt{q_{\mathrm{Re}}^2 + q_{\mathrm{Im}}^2}=\theta_{} r_{\mathrm{d}}$.
    The orange arrows lie within the respective projected plane corresponding to the $xz$-plane and $yz$-plane of the base.
    The virtual displacement can be projected onto the respective plane, resulting in the respective projected virtual displacement $q_{\mathrm{Re}}$ and $q_{\mathrm{Im}}$.
    The corresponding arc parameters $\theta$ and $\phi$ are also shown.
    }
    \label{fig:scheme_segment_manifold}
\end{figure}

The forward mapping $\boldsymbol{f}_{\mathrm{M,II}}$ transforms the Clarke coordinates into arc space parameters.
It leads to
\begin{align}
    \phi_{i}
    &=
    \mathrm{arctan2}(q_{\mathrm{Im},i},q_{\mathrm{Re},i})
    \quad\text{and}
    \label{eq:FM_Manifold2ArcSpace_phi}
    \\
    \theta_{i}
    &=
    \frac{1}{r_{\mathrm{d}}}\sqrt{q_{\mathrm{Re},i}^2 + q_{\mathrm{Im},i}^2},
    \label{eq:FM_Manifold2ArcSpace_theta}
\end{align}
where $\theta_{i}$ is the segment's bending angle and $\phi_{i}$ the segment's bending direction, see Fig.~\ref{fig:scheme_segment_manifold}.
The inverse mapping $\boldsymbol{f}_{\mathrm{M,II}}^{-1}$ can be stated as
\begin{equation}
    \begin{bmatrix}
    q_{\mathrm{Re},i} \\
    q_{\mathrm{Im},i}
    \end{bmatrix} = \theta_{i} r_{\mathrm{d}} \begin{bmatrix}
    \cos({\phi_i}) \\
    \sin({\phi_i})
    \end{bmatrix}.
    \label{eq:IM_Manifold2ArcSpace}
\end{equation}
The physical interpretation of the Clarke coordinates and the corresponding arc parameters are shown in Fig.~\ref{fig:scheme_segment_manifold}
(see~\cite{GrassmannSenykBurgner-Kahrs_submitted_2024} for further details).
By combining the forward and inverse mappings between joint space, manifold, and arc space, a generalized closed-form solution is obtained for the robot-dependent mapping, applicable to an arbitrary number of tendons per segment.

\subsection{Kinetic Energies}
\label{subsec:kinetic_energies}

The total kinetic energy of the \ac{TDCR} is given by
\begin{equation}
    T=T_{\mathrm{b}}+T_{\mathrm{t}_1}+T_{\mathrm{t}_2}+T_{\mathrm{d}}
    .
    \label{eq:T}
\end{equation}
Each component is derived below.

Firstly, the kinetic energy of the backbone comprises translational and rotational components along the segments
\begin{equation}
    \begin{split}
        T_{\mathrm{b}} = 
        \frac{1}{2} 
        \sum_{i=1}^{m}
        \Bigg[ 
        &\int\limits_{0}^{\ell_{i}} 
        \left(
        {}_{(0)}\dot{x}_{s_i}^2 + {}_{(0)}\dot{y}_{s_i}^2 + {}_{(0)}\dot{z}_{s_i}^2
        \right) \rho_{\mathrm{b}}A_{\mathrm{b}} \,\mathrm{d} s_{i} + \\
        & \int\limits_{0}^{\ell_{i}}
        \boldsymbol{\omega}_{s_i}\transpose \boldsymbol{I}_{\mathrm{b}} \boldsymbol{\omega}_{s_i} \rho_{\mathrm{b}} \, \mathrm{d}s_i
        \Bigg]
        ,
    \end{split}
    \label{eq:T_B}
\end{equation}
where $\rho_{\mathrm{b}}$ is the material density of the backbone, $A_{\mathrm{b}}$ represents the cross-sectional area of the backbone and $\boldsymbol{I}_{\mathrm{b}}$ denotes the tensor of the area moments of inertia of the backbone.
Note that ${}_{(0)}\dot{x}_{s_{i}},{}_{(0)}\dot{y}_{s_{i}},{}_{(0)}\dot{z}_{s_{i}}$ are the time derivatives of the cartesian coordinates of a point $\boldsymbol{p}_{s_{i}}$ on the backbone's segment with length $\ell_{i}$.
They are expressed with the basis coordinate frame, denoted by the subscript~$0$, which corresponds to the coordinate frame of the most proximal segment. The coordinate frame of a segment is shown in Fig.~\ref{fig:scheme_segment_manifold}.
The angular velocity $\boldsymbol{\omega}_{s_i} = [\omega_{x_{s_i}}, \omega_{y_{s_i}}, \omega_{z_{s_i}}]\transpose$ describes the rotational motion of the local coordinate frame attached to the backbone at arc length $s_i$.

Secondly, a similar expression can be found for the tendons
\begin{equation}
    \begin{split}
        T_{\mathrm{t}_{1}} = \frac{n}{2} 
        \sum_{i=1}^{m}
        \sum_{j=1}^{i}
        \Bigg[ 
        & \int\limits_{0}^{\ell_{j}} 
        \left(
        {}_{(0)}\dot{x}_{s_{j}}^{2}+
        {}_{(0)}\dot{y}_{s_{j}}^{2}+
        {}_{(0)}\dot{z}_{s_{j}}^{2}
        \right) \rho_{\mathrm{t}}A_{\mathrm{t}} \,\mathrm{d} s_{j}+ \\
        & \int\limits_{0}^{\ell_{i}}
        {\boldsymbol{\omega}_{s_j}\transpose} \boldsymbol{I}_{\mathrm{t}} \boldsymbol{\omega}_{s_j} \rho_{\mathrm{t}} \, \mathrm{d}s_j
        \Bigg]
        ,
    \end{split}
    \label{eq:T_T1}
\end{equation}
where subscript $\mathrm{t}$ refers to the parameters of the tendons.
Note that the entire length of the tendons must be taken into account, as the tendons of the distal segments are also routed through the proximal segments.

The change of tendon displacements causes a further component of the kinetic energy
\begin{equation}
    T_{\mathrm{t}_{2}}=\sum_{i=1}^{m}  \sum_{k=1}^{n} \frac{1}{2} \rho_{\mathrm{t}}A_{\mathrm{t}} ({\ell_i}+{q}_{ik}) \dot{q}_{ik}^{2},
    \label{eq:T_T2}
\end{equation}
with the change of tendon displacement over time of the tendons of segment $i$
\begin{equation}
    \begin{split}
        \dot{\boldsymbol{q}}_{i}
        =
        \frac{\mathrm{d}}{\mathrm{d}t}
        \sum_{j=1}^{i} \MPinv
        \begin{bmatrix}
        q_{\mathrm{Re},j}\\
        q_{\mathrm{Im},j}
        \end{bmatrix}
        ,
    \end{split}
\end{equation}
with $i = 1,\dots,m$.

Thirdly, each spacer disk is fixed perpendicularly to the backbone.
The kinetic energy of all spacer disks can be determined as
\begin{equation}
    \begin{split}
    T_{\mathrm{d}}=\frac{1}{2} 
    \sum_{i=1}^{m}
    \Bigg[
    & \sum_{o=1}^{D_{i}} 
    \left(
    {}_{(0)}\dot{x}_{\mathrm{d}_{o}}^{2}+
    {}_{(0)}\dot{y}_{\mathrm{d}_{o}}^{2}+
    {}_{(0)}\dot{z}_{\mathrm{d}_{o}}^{2}
    \right) 
    m_{\mathrm{d}} + \\
    & \sum_{o=1}^{D_{i}}
    \boldsymbol{\omega}_{\mathrm{d}_{o}}\transpose \boldsymbol{I}_{\mathrm{d}} \boldsymbol{\omega}_{\mathrm{d}_{o}}
    \Bigg]
    ,
    \end{split}
    \label{eq:T_D}
\end{equation}
where subscript $\mathrm{d}$ indicates the parameters of a spacer disk.
The linear and angular velocities of each spacer disks can be derived by substituting $s_{i}=o \cdot h_{i}$ for $o=1,2,\dots,D_{i}$ with the total number of spacer disks of the $i$-th segment $D_{i}=\ell_{i}/h_{i}$ and the distance between each spacer disk $h_{i}$.

\subsection{Potential Energies}
\label{subsec:potential_energies}

The total potential energy can be expressed as
\begin{equation}
    U=U_{\mathrm{g,b}}+U_{\mathrm{g,t}}+U_{\mathrm{g,d}}+U_{\mathrm{e,b}}
    .
\end{equation}
The four contributions to $U$ are defined in the following.

Firstly, supposing that the direction of the $-z$-axis of the \ac{TDCR}'s base coordinate frame coincides with the direction of the gravitational vector with acceleration $g$, the gravitational potential energy of the backbone can be represented by
\begin{equation}
    U_{\mathrm{g,b}}=\sum_{i=1}^{m}\int\limits_{0}^{\ell_{i}} {}_{(0)}z_{s_{i}} \rho_{\mathrm{b}} A_{\mathrm{b}} g \,\mathrm{d} s_{i}.
    \label{eq:U_g,B}
\end{equation}
The gravitational potential energy of the tendons is given by
\begin{equation}
    U_{\mathrm{g,t}}=
    n
    \sum_{i=1}^{m}
    \sum_{j=1}^{i}
    \int\limits_{0}^{\ell_{j}} {}_{(0)}z_{s_{j}} \rho_{\mathrm{t}} A_{\mathrm{t}} g \,\mathrm{d} s_{j},
\label{eq:U_g,T}
\end{equation}
whereas the potential energy of all spacer disks is given by
\begin{equation}
    U_{\mathrm{g,d}}=\sum_{i=1}^{m}
    \sum_{o=1}^{D_{i}} {}_{(0)}z_{\mathrm{d}_{o}} m_{\mathrm{d}} g 
    .
    \label{eq:U_g,D}
\end{equation}

Secondly, according to the Euler-Bernoulli beam theory~\cite{Bauchau2009}, the elastic potential energy is
\begin{align}
    U_{\mathrm{e,b}} = 
    & \sum_{i=1}^{m}
    \int\limits_{0}^{\ell_{i}}
    \left(
    \left(
    \frac{E_{\mathrm{b}}{I}_{\mathrm{b},zz}}{2}  +
    n \frac{E_{\mathrm{t}}{I}_{\mathrm{t},zz}}{2}
    \right) \left(\frac{\mathrm{d} \theta_{s_{i}}}{\mathrm{d}s_{i}}\right)^2
    \right)
    \,\mathrm{d}s_{i} 
    \nonumber\\
    = 
    & \sum_{i=1}^{m} 
    \left(
    \frac{E_{\mathrm{b}}{I}_{\mathrm{b},zz}}{2\ell_{i}} \theta_{i}^2+
    n \frac{E_{\mathrm{t}}{I}_{\mathrm{t},zz}}{2\ell_{i}} \theta_{i}^2
    \right)
    ,
\label{eq:U_e,B}
\end{align}
where ${I}_{\mathrm{b},zz}$ and ${I}_{\mathrm{t},zz}$ denotes the area second moments about the segments' $z$-axis.

\subsection{Damping Forces}
\label{subsec:damping_forces}

To account for energy dissipation, we introduce damping into the dynamic model.
Assuming homogeneous material properties for the backbone, we employ a linear damping model analogous to~\cite{Katzschmann2019}.

The damping model for segment $i$ in arc space results in
\begin{equation}
    \boldsymbol{D}_{\mathrm{a},i}=
    \begin{bmatrix}
    d_{\theta,i}\theta_{i}^{2} & 0 \\
    0 & d_{\theta,i}
    \end{bmatrix},
    \label{eq:damping_matrix}
\end{equation}
with the damping coefficient $d_{\theta,i}$.
The damping matrix $\boldsymbol{D}_{\mathrm{a}}\in\mathbb{R}^{2m\times2m}$ of the entire \ac{TDCR} are block diagonal concatenations of $\boldsymbol{D}_{\mathrm{a},i}$.

\subsection{Equation of Motion}
\label{subsec:equation_of_motion}

We can state the \ac{TDCR}'s equation of motion using the Euler-Lagrange formalism

\begin{equation}
     \frac{\mathrm{d}}{\mathrm{d}t}\left(\frac{\partial T}{\partial \dot{\boldsymbol{a}}}\right)-\frac{\partial T}{\partial \boldsymbol{a}}+\frac{\partial U}{\partial \boldsymbol{a}} = \boldsymbol{\tau}_{\mathrm{a}},
\end{equation}
where $\boldsymbol{a}=[ \phi_{1},\theta_{1},\dots,\phi_{m},\theta_{m} ]\transpose\in\mathbb{R}^{2m}$ represents the \acp{TDCR} arc space parameters and $\dot{\boldsymbol{a}}\in\mathbb{R}^{2m}$ and $\ddot{\boldsymbol{a}}\in\mathbb{R}^{2m}$ their first and second time derivatives.
We first derived the dynamic equations in arc space and mapped them onto the manifold, as described in the following.
Note that one might define the energies on the manifold using \eqref{eq:FM_Manifold2ArcSpace_phi}, \eqref{eq:FM_Manifold2ArcSpace_theta}, and their time derivatives.

The equation of motion in arc space notation with damping forces results in
\begin{equation}
    \boldsymbol{M}_{\mathrm{a}}(\boldsymbol{a})\ddot{\boldsymbol{a}}+
    \boldsymbol{C}_{\mathrm{a}}(\boldsymbol{a},\dot{\boldsymbol{a}})\dot{\boldsymbol{a}}+\boldsymbol{g}_{\mathrm{a}}(\boldsymbol{a})+\boldsymbol{K}_{\mathrm{a}}\boldsymbol{a}+\boldsymbol{D}_{\mathrm{a}}(\boldsymbol{a})\dot{\boldsymbol{a}}
    =
    \boldsymbol{\tau}_{\mathrm{a}}
    \label{eq:DynamicModel_ArcSpace_wDamping}
\end{equation}
where $\boldsymbol{M}_{\mathrm{a}}\in\mathbb{R}^{2m\times2m}$ is the mass matrix, $\boldsymbol{C}_{\mathrm{a}}(\boldsymbol{a},\dot{\boldsymbol{a}})\in\mathbb{R}^{2m\times2m}$ contains the Coriolis and centrifugal expressions, $\boldsymbol{g}_{\mathrm{a}}(\boldsymbol{a})\in\mathbb{R}^{2m}$ represents the gravitational forces and $\boldsymbol{K}_{\mathrm{a}}\in\mathbb{R}^{2m\times2m}$ the linear elastic field.
Vector $\boldsymbol{\tau}_{\mathrm{a}}\in\mathbb{R}^{2m}$ are the generalized forces acting on the arc space parameters.

To transform the dynamic model \eqref{eq:DynamicModel_ArcSpace_wDamping} into the manifold space, we use $\boldsymbol{a}$, $\dot{\boldsymbol{a}}$, and $\ddot{\boldsymbol{a}}$ w.r.t. $\boldsymbol{q}_{\mathrm{M}}$, $\dot{\boldsymbol{q}}_{\mathrm{M}}$, and $\ddot{\boldsymbol{q}}_{\mathrm{M}}$, respectively.
They are defined by
\begin{equation}
    \left.
    \begin{aligned}
    \boldsymbol{a}=&\boldsymbol{f}_{\mathrm{M,II}}(\boldsymbol{q}_{\mathrm{M}}) \\
    \dot{\boldsymbol{a}}=&{}^{\mathrm{a}}\boldsymbol{J}_{\mathrm{M}}\dot{\boldsymbol{q}}_{\mathrm{M}} \\
    \ddot{\boldsymbol{a}}=&{}^{\mathrm{a}}\dot{\boldsymbol{J}}_{\mathrm{M}}\dot{\boldsymbol{q}}_{\mathrm{M}}+{}^{\mathrm{a}}\boldsymbol{J}_{\mathrm{M}}\ddot{\boldsymbol{q}}_{\mathrm{M}}
    \end{aligned}
    \right\}
    \label{eq:arc_space_variable_wrt_manifold_space_variables}
\end{equation}
where $\boldsymbol{q}_{\mathrm{M}}=[ q_{\mathrm{Re},1},q_{\mathrm{Im},1},\dots,q_{\mathrm{Re},m},q_{\mathrm{Im},m}]\transpose\in\mathbb{R}^{2m}$ represents the manifold parameters and $\dot{\boldsymbol{q}}_{\mathrm{M}}\in\mathbb{R}^{2m}$ and $\ddot{\boldsymbol{q}}_{\mathrm{M}}\in\mathbb{R}^{2m}$ is their first and second time derivative, respectively.
Here, ${}^{\mathrm{a}}\boldsymbol{J}_{\mathrm{M}}:\mathbb{R}^{2m}\to\mathbb{R}^{2m}$ denotes the Jacobian of $\boldsymbol{f}_{\mathrm{M,II}}(\boldsymbol{q}_{\mathrm{M}})$, \textit{i.e.} $\frac{\partial\boldsymbol{f}_{\mathrm{M,II}}(\boldsymbol{q}_{\mathrm{M}})}{\partial\boldsymbol{q}_{\mathrm{M}}}$, while ${}^{\mathrm{a}}\dot{\boldsymbol{J}}_{\mathrm{M}}$ represents its time derivative.
The generalized forces $\boldsymbol{\tau}_{\mathrm{a}}$ are transformed using
\begin{equation}
    \boldsymbol{\tau}_{\mathrm{M}}={}^{\mathrm{a}}\boldsymbol{J}_{\mathrm{M}}\transpose \boldsymbol{\tau}_{\mathrm{a}}.
    \label{eq:transformation_arc_space_force_into_manifold_forces}
\end{equation}

Inserting~(\ref{eq:arc_space_variable_wrt_manifold_space_variables}) and~(\ref{eq:transformation_arc_space_force_into_manifold_forces}) into~(\ref{eq:DynamicModel_ArcSpace_wDamping}) yields the manifold model
\begin{equation}
    \begin{split}
        \boldsymbol{M}_{\mathrm{M}}(\boldsymbol{q}_{\mathrm{M}})\ddot{\boldsymbol{q}}_{\mathrm{M}}
        +
        \boldsymbol{C}_{\mathrm{M}}(\boldsymbol{q}_{\mathrm{M}},\dot{\boldsymbol{q}}_{\mathrm{M}})
        \dot{\boldsymbol{q}}_{\mathrm{M}}+
       \\
        \boldsymbol{g}_{\mathrm{M}}(\boldsymbol{q}_{\mathrm{M}})
        +
        \boldsymbol{K}_{\mathrm{M}}\boldsymbol{q}_{\mathrm{M}}+
        \boldsymbol{D}_{\mathrm{M}}\dot{\boldsymbol{q}}_{\mathrm{M}}
        =
        \boldsymbol{\tau}_{\mathrm{M}}
        ,
    \end{split}
    \label{eq:DynamicModel_Manifold_wDamping}
\end{equation}
where the components are defined as
\begin{equation}
    \begin{aligned}
    \boldsymbol{M}_{\mathrm{M}}(\boldsymbol{q}_{\mathrm{M}}) =& {}^{\mathrm{a}}\boldsymbol{J}_{\mathrm{M}}\transpose\boldsymbol{M}_{\mathrm{a}}(\boldsymbol{f}_{\mathrm{M,II}}(\boldsymbol{q}_{\mathrm{M}})){}^{\mathrm{a}}\boldsymbol{J}_{\mathrm{M}}
    \\
    \boldsymbol{C}_{\mathrm{M}}(\boldsymbol{q}_{\mathrm{M}},\dot{\boldsymbol{q}}_{\mathrm{M}}) =&
        {}^{\mathrm{a}}\boldsymbol{J}_{\mathrm{M}}\transpose\boldsymbol{M}_{\mathrm{a}}(\boldsymbol{f}_{\mathrm{M,II}}(\boldsymbol{q}_{\mathrm{M}})){}^{\mathrm{a}}\dot{\boldsymbol{J}}_{\mathrm{M}} + \\
        & {}^{\mathrm{a}}\boldsymbol{J}_{\mathrm{M}}\transpose\boldsymbol{C}_{\mathrm{a}}(\boldsymbol{f}_{\mathrm{M,II}}(\boldsymbol{q}_{\mathrm{M}}),{}^{\mathrm{a}}\boldsymbol{J}_{\mathrm{M}}\dot{\boldsymbol{q}}_{\mathrm{M}}){}^{\mathrm{a}}\boldsymbol{J}_{\mathrm{M}}
        \\
    \boldsymbol{g}_{\mathrm{M}}(\boldsymbol{q}_{\mathrm{M}}) =& {}^{\mathrm{a}}\boldsymbol{J}_{\mathrm{M}}\transpose\boldsymbol{g}_{\mathrm{a}}(\boldsymbol{f}_{\mathrm{M,II}}(\boldsymbol{q}_{\mathrm{M}}))
    \\
    \boldsymbol{K}_{\mathrm{M}}\boldsymbol{q}_{\mathrm{M}} =& {}^{\mathrm{a}}\boldsymbol{J}_{\mathrm{M}}\transpose\boldsymbol{K}_{\mathrm{a}}\boldsymbol{f}_{\mathrm{M,II}}(\boldsymbol{q}_{\mathrm{M}})
    \\
    \boldsymbol{D}_{\mathrm{M}} =& {}^{\mathrm{a}}\boldsymbol{J}_{\mathrm{M}}\transpose\boldsymbol{D}_{\mathrm{a}}(\boldsymbol{f}_{\mathrm{M,II}}(\boldsymbol{q}_{\mathrm{M}})){}^{\mathrm{a}}\boldsymbol{J}_{\mathrm{M}}
    \end{aligned}
    \nonumber
\end{equation}
Note that the stiffness matrix $\boldsymbol{K}_{\mathrm{M}}$ and damping matrix $\boldsymbol{D}_{\mathrm{M}}$ are constant on the manifold, with $\boldsymbol{D}_{\mathrm{M}}$ containing only the segment's damping coefficients $d_{\theta,i}$ on its main diagonal~\cite{Santina2020}, while $\boldsymbol{K}_{\mathrm{M}}$ is derived from elastic potential energy.
Additionally, no external forces act on the \ac{TDCR}.

\subsection{Generalized Forces on Manifold}
\label{subsec:generalized_forces_on_manifold}

To map the tendon forces from joint space to manifold, we use the closed-form Jacobian ${}^{\mathrm{M}}\boldsymbol{J}_{\mathrm{q}}:\mathbb{R}^{m \times n}\to\mathbb{R}^{2m}$ of $\boldsymbol{f}_{\mathrm{M,I}}(\boldsymbol{q}_{})$, \textit{i.e.} $\frac{\partial\boldsymbol{f}_{\mathrm{M,I}}(\boldsymbol{q}_{})}{\partial\boldsymbol{q}_{}}$. 
Its transpose corresponds structurally to ${}^{\mathrm{M}}\boldsymbol{J}_{\mathrm{q}}\transpose=\frac{2}{n}\MPinv$.
The generalized forces $\boldsymbol{\tau}_{\mathrm{M},i}$ are transformed using

\begin{equation}
    \boldsymbol{F}_{i}=
    {}^{\mathrm{M}}\boldsymbol{J}_{\mathrm{q}}\transpose
    \boldsymbol{\tau}_{\mathrm{M},i}
    =
    \frac{2}{n}\MPinv
    \boldsymbol{\tau}_{\mathrm{M},i}
    .
    \label{eq:IM_Torques_JointSpace2Manifold}
\end{equation}

As $\MP\MPinv = \Imat_{2 \times 2}$ according to \cite{GrassmannSenykBurgner-Kahrs_submitted_2024}, multiplying \eqref{eq:IM_Torques_JointSpace2Manifold} on the left by $\frac{n}{2}\MP$ yields the inverse mapping
\begin{equation}
    \boldsymbol{\tau}_{\mathrm{M},i}
    =
    \frac{n}{2}\MP
    \boldsymbol{F}_{i},
    \label{eq:FM_Torques_JointSpace2Manifold}
\end{equation}
where vector $\boldsymbol{F}_{i}\in\mathbb{R}^{\mathrm{n}}$ represents the tendon forces.

%% file: Control_Strategies/control_strategies.tex
\section{Control Strategies}
\label{sec:control_strategies}

\begin{figure*}[tbp]
    \vspace*{0.75em}
    \centering
    \includegraphics[width=0.9\textwidth]{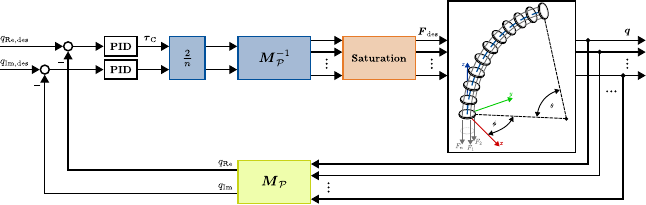}
    \vspace{-0.75em}
    \caption{
    Block diagram of the proposed controllers operating on the manifold:
    The desired trajectory, $q_{\mathrm{Re,des}}$  and  $q_{\mathrm{Im,des}}$, serves as input to the controller, which outputs the generalized forces on the manifold $\boldsymbol{\tau}_{\mathrm{C}}$.
    These are mapped into joint space using the blue blocks, which implement~(\ref{eq:IM_Torques_JointSpace2Manifold}).
    The orange block represents a saturation step, which prevents negative tendon forces by applying one of the proposed methods—clipping, redistribution, or shifting—as detailed in Sec.~\ref{sec:control_strategies}.
    The measured tendon displacements \(\boldsymbol{q}_{}\) are transformed onto the manifold using the green block, which follows~(\ref{eq:FM_JointSpace2Manifold}).
    The diagram depicts the PID controller, whereas the PD controller omits the integral component in the controller gains.
    }
    \label{fig:controlscheme_new}
    \vspace{-1.5em}
\end{figure*}

Controls in joint space typically require a separate controller for each joint, where the outputs of all controllers are subject to \eqref{eq:tendon_constraint}.
The Clarke transform, on the other hand, allows for a constraint-informed controller by utilizing a linear mapping from $n$ tendon displacements to the $\SI{2}{\ac{DoF}}$ manifold.
Therefore, it enables control exclusively on the manifold using only two control parameters per segment.
Subsequently, we implement linear PID and PD controllers.
As depicted in~Fig.~\ref{fig:controlscheme_new}, the output of the controllers are the generalized forces on the manifold, which are transformed into the joint space using~(\ref{eq:IM_Torques_JointSpace2Manifold}).
This may result in negative tendon forces, which are physically infeasible.
To overcome the negative tendon forces, we propose three strategies: clipping, redistribution, and shifting.

\subsection{Clipping}
\label{subsubsec:clipping_forces}

This strategy is straightforward, and it clips negative tendon forces to zero.
However, it modifies the generalized forces on the manifold when transforming back from joint space, so that they no longer match the controller’s output.
While we observe that this method can maintain stability for linear controllers, it can negatively impact control performance, as shown in Sec.~\ref{subsec:simulative_validation}.

\subsection{Redistributing}
\label{subsubsec:redistribute_forces}

This strategy redistributes all tendon forces to tendons with only positive tendon force by projecting the tendon forces onto the segment’s bending direction.
Focusing the redistribution to the two tendons closest to the bending direction ensures an analytical solution, as this yields a determined system of equations.
While this method maintains consistent generalized forces on the manifold, its implementation is more complex than the shifting method, which provides equivalent solution quality.
For readability, the redistribution approach is not further detailed here.

\subsection{Shifting}
\label{subsubsec:shift_forces}

Shifting the tendon forces by the segment’s smallest force, defined as ${F}_{\mathrm{min},i}=\min({\boldsymbol{{F}}_{i}})$, ensures that all tendon forces remain non-negative.
The back transformation preserves the original generalized forces on the manifold
\begin{equation}
    \begin{split}
    \boldsymbol{\tau}_{M,i}
    &=
    \frac{n}{2}{{\MP}_{}} \boldsymbol{{F}}_{\mathrm{shifted},i}\\
    &=
    \frac{n}{2}{{\MP}_{}} \left(\boldsymbol{{F}}_{\mathrm{init},i} + {F}_{\mathrm{min},i}\boldsymbol{1}_{n}\right)\\
    &=
    \frac{n}{2}{{\MP}_{}} \boldsymbol{{F}}_{\mathrm{init},i},
    \end{split}
    \label{eq:transform_shifted_force_vector}
\end{equation}
where $\boldsymbol{1}_{n}$ is a vector with $n$ ones.
The additional term $\frac{n}{2}{{\MP}_{}} {F}_{\mathrm{min},i}\boldsymbol{1}_{n}$ simplifies to zero because $\sum_{i=1}^{n}{\cos({\psi_{i}})}=0$ and $\sum_{i=1}^{n}{\sin({\psi_{i}})}=0$ hold \cite{GrassmannSenykBurgner-Kahrs_submitted_2024}.

We note that shifting increases tendon tension and, therefore, it might change the stiffness of a \ac{TDCR} \cite{Butler2019, PeyronBurgner-Kahrs_TRO_2023}.
Furthermore, pretension relates to static frictional forces \cite{Butler2019}.
Thus, this may indicate that static friction could be reduced by adding more tendons that distribute the pretension more evenly.

%% file: Validation/validation.tex
\section{Validation}
\label{sec:validation}

We validate our proposed dynamic model and control schemes through both simulations and experiments.

\subsection{\ac{TDCR} Parameters}
\label{subsec:experimental_setup}

We build a \ac{TDCR} prototype depicted in Fig.~\ref{fig:test_bench} with one segment and five tendons for the experimental evaluation, where we make use of the OpenCR project~\cite{Grassmann2024}.
The backbone is a Nitinol rod characterized by its material density $\rho_{\mathrm{b}} = \SI{6400}{kg/m^3}$, elastic modulus $E_{\mathrm{b}} = \SI{58}{MPa}$, diameter $d_{\mathrm{b}} = \SI{1}{mm}$, and length $l = \SI{0.2}{m}$.
Ten spacer disks are equally distributed along the backbone, where each has a mass of $m_{\mathrm{d}} = \SI{0.81}{g}$.
Five tendons are equally distributed around the circumference of a circle with a radius $r_{\mathrm{d}} = \SI{7}{mm}$.
We also use a quasi-direct-drive actuation unit~\cite{Grassmann2024} for each tendon.
Its proprioception allows for real-time tendon-tension control.

For the simulation, a \ac{TDCR} with two segments is implemented in \textsc{Matlab} 2023b using the Dormand-Prince algorithm for numerical integration.
The material and geometric parameters match those of the experimental setup.
Additionally, the damping coefficient $d_{\theta,i}=\SI{11.27e-4}{\newton\meter\second}$ for \eqref{eq:damping_matrix} is experimentally identified assuming uniform damping for each segment and using regression with a pseudo-inverse.

\subsection{Simplifying Dynamic Model}
\label{subsec:influence_translational_rotational_forces}

\begin{figure}[tbp]
    \centering
    \includegraphics[width=\columnwidth]{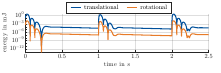} 
    \vspace*{-2.5em}
    \caption{Translational and rotational energies of one segment during a step trajectory where the segment is controlled to bend to $\theta=\frac{\pi}{4}$, then rotate to $\phi=\frac{\pi}{4}$ while maintaining the bending, and finally return to its initial straight position. Note that the energy is scaled logarithmically.}
    \label{fig:comp_trans_rot_energies}
    \vspace{-0.75em}
\end{figure}

We compare the total kinetic energies~\eqref{eq:T} with the rotational energies of the system in \eqref{eq:T_B}, \eqref{eq:T_T1}, and \eqref{eq:T_D}. 
Figure~\ref{fig:comp_trans_rot_energies} shows that the translational kinetic energy dominates the rotational energy.
The proximal rotational movements lead to translational movements in distal sections~\cite{Falkenhahn2015}. 
Additionally, the slender shape of \acp{TDCR} results in a low moment of inertia compared to the backbone’s mass, emphasizing translational over rotational energy.
Therefore, we neglect the rotational terms in \eqref{eq:T_B}, \eqref{eq:T_T1}, and \eqref{eq:T_D}.

\begin{figure}[tbp]
    \centering
    \includegraphics[width=\columnwidth]{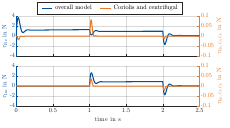}
    \vspace*{-2.5em}
    \caption{Overall forces and Coriolis and centrifugal forces of one segment during a step trajectory where the segment is controlled to bend to $\theta=\frac{\pi}{4}$, then rotate to $\phi=\frac{\pi}{4}$ while maintaining the bending, and finally return to its initial straight position. Note that the left ordinate applies to overall forces, while the right ordinate applies to the Coriolis and centrifugal forces.}
    \label{fig:comp_total_coriolis_forces}
    \vspace{-1.5em}
\end{figure}

Furthermore, the Coriolis and centrifugal forces are compared with the total forces working in the system, see Fig.~\ref{fig:comp_total_coriolis_forces}.
Coriolis and centrifugal forces are also negligible hereinafter due to low velocities~\cite{Falkenhahn2015}, and due to low masses in our system.

Additionally, we omit the kinetic and potential energies of the tendons, \textit{i.e.}, \eqref{eq:T_T1}, \eqref{eq:T_T2}, and \eqref{eq:U_g,T}.
In our experimental setup, the weight of the tendons is assumed to be negligible compared to that of the spacer disks, as their contribution to the total system mass is minimal.

\subsection{Validation in Simulation}
\label{subsec:simulative_validation}

The model in simulation accounts for $2^{m}-1$ critical configurations that arise when $m$ individual segments approach their straight configurations.
While the dynamic equations on the manifold are analytically well-conditioned~\cite{Santina2020}, numerical instabilities occur in these cases.
To address this, symbolic limit expressions are derived for all such cases and are automatically applied during the simulation if the generalized coordinates fall below a threshold value of $\{q_{\mathrm{Re},i},q_{\mathrm{Im},i}\}=\SI{5e-6}{m}$.

The system is controlled using a PID controller, with initial configurations set to the straight position $[q_{\mathrm{Re},1},q_{\mathrm{Im},1},q_{\mathrm{Re},2},q_{\mathrm{Im},2}]\transpose=[0,0,0,0]\transpose\SI{}{\meter}$.
Sine wave trajectories are defined with initial frequencies of $[f_{\mathrm{Re},1},f_{\mathrm{Im},1},f_{\mathrm{Re},2},f_{\mathrm{Im},2}]\transpose = [0.1,0.05,0.15,0.2]\transpose\SI{}{Hz}$ and amplitudes of $[q_{\text{amp,Re},1},q_{\text{amp,Im},1},q_{\text{amp,Re},2},q_{\text{amp,Im},2}]\transpose = [0.01,0.005,0.005,0.025]\transpose\SI{}{\meter}$.
The frequencies increase linearly by $\SI{0.005}{Hz/s}$ over time to test the system’s tracking performance under dynamic changes.
These trajectories are chosen as they capture a range of relevant dynamic effects, including different bending and oscillation frequencies.

\begin{figure*}[tbp]
    \vspace*{0.75em}
    \centering
    \includegraphics[width=\textwidth]{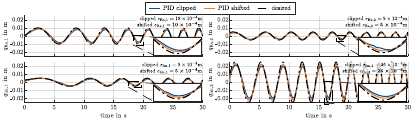}
    \vspace{-2.25em}
    \caption{Simulation results for tracking of desired sine wave trajectories on the manifold with increasing frequencies with a PID controller.
    The shifting and clipping methods are used to prevent negative tendon forces.
    Detailed views around the trajectory peaks highlight differences between the methods.
    The \ac{RMSE} $\{e_{\mathrm{Re},i},e_{\mathrm{Im},i}\}$ is shown for performance evaluation.
    }
    \label{fig:control_simulation}
    \vspace{-1.5em}
\end{figure*}

The system's tracking performance on the manifold is illustrated in Fig.~\ref{fig:control_simulation}.
The controller gains are $\mathbf{K}_p = \SI{1500}{\frac{\newton}{\metre}}$, $\mathbf{K}_i = \SI{1500}{\frac{\newton}{\metre\second}}$, and $\mathbf{K}_d = \SI{1}{\frac{\newton\second}{\metre}}$.
The force adaptation methods of clipping and shifting are tested to prevent negative tendon forces.
The shifting approach ensures unchanged torques on the manifold and outperforms the clipping approach, achieving an average $\SI{43.3}{\percent}$ lower \ac{RMSE} across both segments compared to clipping.

\subsection{Experimental Validation}
\label{subsec:experimental_validation}

The experimental validation evaluates the system’s ability to track a dynamic trajectory on the manifold.
The desired trajectory follows a bending motion in the segments' $xz$-plane with an initial frequency of \SI{0.1}{Hz} and amplitudes $[q_{\text{amp,Re},1},q_{\text{amp,Im},1}]\transpose = [0.01,0.0]\transpose\SI{}{m}$.
The frequency increases linearly at a rate of \SI{0.005}{Hz/s}, testing the controller’s performance under dynamically changing conditions.

\begin{figure}[tbp]
    \centering
    \includegraphics[width=\columnwidth]{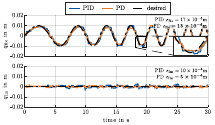}
    \vspace{-2.25em}
    \caption{Experimental results for position control of a desired sine wave trajectory on the manifold using a PID and PD controller with the proposed shifting method.
    The desired trajectory corresponds to a bending motion in the $xz$-plane of the \ac{TDCR} segment.
    A detailed view around the trajectory peaks highlights differences between the methods.
    The \ac{RMSE} $\{e_{\mathrm{Re}},e_{\mathrm{Im}}\}$ is shown for performance evaluation.
    }
    \label{fig:control_testbench}
    \vspace{-1.5em}
\end{figure}

Two controllers are evaluated, each utilizing the shifting method: a PID controller with anti-windup for the integrative term and a PD controller.
The PID controller uses $\mathbf{K}_p = \SI{1000}{\frac{\newton}{\metre}}$, $\mathbf{K}_i = \SI{1000}{\frac{\newton}{\metre\second}}$, and $\mathbf{K}_d = \SI{0.25}{\frac{\newton\second}{\metre}}$, with an anti-windup saturation limit of \SI{0.2}{\newton}, while the PD controller uses $\mathbf{K}_p = \SI{1750}{\frac{\newton}{\metre}}$ and $\mathbf{K}_d = \SI{0.55}{\frac{\newton\second}{\metre}}$.
The experimental results of the position controllers are shown in Fig.~\ref{fig:control_testbench}.  
The trajectory is tracked exclusively on the manifold, confirming the practical feasibility of manifold-based control in a real robotic system. 
Since the Clarke coordinates cannot be directly observed, they are computed by transforming the measured tendon displacements $\boldsymbol{q}$ onto the manifold using~(\ref{eq:FM_JointSpace2Manifold}).  
This enables direct comparison with the desired trajectory on the manifold, see Fig.~\ref{fig:controlscheme_new}.
The PID controller tracks the trajectory robustly but is influenced by ripple effects caused by cogging torque from the actuator units.  
The PD controller achieves smoother trajectories while following the desired trajectory, achieving an average \SI{37.1}{\percent} lower \ac{RMSE} compared to the PID controller.

%% file: Discussion/discussion.tex
\section{Discussion}
\label{sec:discussion}

In this section, we interpret the results of the proposed dynamic model and control strategies.

\subsection{Model Formulation}
\label{subsec:model_formulation}

A key advantage over alternative formulations is its representation on a \SI{2}{\ac{DoF}} manifold.
By leveraging the Clarke transform~\cite{GrassmannSenykBurgner-Kahrs_submitted_2024}, our dynamic modeling approach inherently satisfies tendon constraints~(\ref{eq:tendon_constraint}).
A constraint-informed approach eliminates the need for additional constraint-handling and provides a computationally efficient framework~\cite{GrassmannSenykBurgner-Kahrs_submitted_2024}.
While previous methods have primarily been introduced for up to four tendons per segment~\cite{Santina2020}, our approach generalizes to \acp{TDCR} with $n\geq 3$ tendons per segment.
Consequently, the proposed model broadens the design space without increasing computational overhead.

The assumption of uniform mass distribution along the backbone leads to integral-based formulations of the kinetic and potential energies over each segment’s length, see (\ref{eq:T_B}), (\ref{eq:T_T1}), (\ref{eq:U_g,B}), and (\ref{eq:U_g,T}).
This results in complex symbolic expressions~\cite{Godage2015}, especially for multiple segments.
However, simulation results indicate a negligible impact of rotational energies~(Fig.~\ref{fig:comp_trans_rot_energies}) and Coriolis and centrifugal forces~(Fig.~\ref{fig:comp_total_coriolis_forces}), allowing for their exclusion.
Consequently, we can reduce the symbolic complexity without significantly affecting accuracy, which is in accordance with~\cite{Falkenhahn2015}.

Although the dynamic model on the \SI{2}{\ac{DoF}} manifold is analytically singularity-free~\cite{Santina2020}, we observe numerical instabilities when \ac{TDCR} segments approach their straight configuration.
These instabilities arise from poor conditioning in the numerical integration routine.
To address this, we use symbolic limit models, which are automatically applied when the segments approach a straight configuration. 
While this method successfully enables continuous simulations without bypassing critical configurations, it poses computational challenges for multi-segment \acp{TDCR} in \textsc{Matlab}’s symbolic engine.
Consequently, we rely on small-value substitutions, which may introduce control instabilities, such as oscillatory behavior in the segments.
Nevertheless, these approaches ensure stable simulation and model continuity, even near straight configurations.
However, a comprehensive simulation study, covering complex deformations, high load-to-stiffness ratios, and benchmarking against alternative models, remains subject of future work.

\subsection{Control Results}
\label{subsec:control_results}

The presented simulations evaluate different strategies for handling negative tendon forces.
While redistribution maintains the generalized forces on the manifold, it concentrates all tendon forces on only two tendons~(Sec.~\ref{subsubsec:shift_forces}), counteracting the benefit of distributing actuation forces across more tendons~\cite{Butler2019}, and thus is not considered further in our simulations.
The shifting approach maintains consistent generalized forces on the manifold~(\ref{eq:transform_shifted_force_vector}) as well, preventing physically infeasible forces without introducing instabilities in control~(Fig.~\ref{fig:control_simulation}).
Moreover, by increasing the minimum possible offset $F_{\mathrm{min},i}$ in~(\ref{eq:transform_shifted_force_vector}), shifting also enables selective stiffness adjustments of a \ac{TDCR} segment.
This effect is achieved by modifying tendon pretension~\cite{Butler2019}.  
In future work, the shifting method could be leveraged for stiffness tuning in real-world applications without compromising position control performance.

The simulations confirm that the proposed linear control strategies enable stable and accurate trajectory tracking for \acp{TDCR} with $n$ tendons per segment and multiple segments.
Each segment can be controlled with two key variables (Fig.~\ref{fig:controlscheme_new}), which inherently satisfies tendon constraints~(\ref{eq:tendon_constraint}) and avoids the need for complex coordination between tendons.

In our experiments, the linear controllers achieve real-time control at \SI{1}{\kilo\hertz}~\cite{Grassmann2024}.
The PID and PD controllers demonstrate robust trajectory tracking, though cogging effects in the actuator units cause step-like motion, especially at low speeds and low actuation forces~(Fig.~\ref{fig:control_testbench}).
The PID controller requires an anti-windup strategy to prevent integrator windup, which would otherwise lead to slow response times and poor tracking accuracy.
The PD controller achieves smoother trajectories but introduces increasing errors with larger bending, as the elastic backbone stores more energy~\cite{Grassmann2024}. 
Future work should address dynamic modeling and compensation of cogging torque to improve control performance, and assess controller robustness under model uncertainties and external disturbances to further validate practical applicability.

%% file: Conclusion/conclusion.tex
\section{Conclusion}
\label{sec:conclusion}

We present a computationally efficient and generalized dynamic model for \acp{TDCR} with multiple segments and an arbitrary number of tendons per segment.
Our proposed model is based on the Clarke transform, the Euler-Lagrange formalism, and the \ac{CC} assumption.
This model allows us to synthesize constraint-informed controllers.